\documentclass[lettersize,journal]{IEEEtran}
\usepackage{amsmath,amssymb,amsfonts}
\usepackage{algorithmicx,algorithm}
\usepackage{amsthm,amsmath,amssymb}
\usepackage{mathrsfs}
\usepackage{array}
\usepackage{textcomp}
\usepackage{stfloats}
\usepackage{verbatim}
\usepackage{graphics} 
\usepackage{epsfig} 
\usepackage{times} 
\usepackage{amsmath} 
\usepackage{amssymb}  
\usepackage{cite} 
\usepackage{graphicx}
\usepackage{subfigure} 
\usepackage{picinpar}
\usepackage{url}
\usepackage{flushend}
\usepackage{colortbl}
\usepackage{soul}
\usepackage{multirow}
\usepackage{pifont}
\usepackage{color}
\usepackage{alltt}
\usepackage{hyperref}
\usepackage{float}
\usepackage{enumerate}
\usepackage{siunitx}
\usepackage{breakurl}
\usepackage{epstopdf}
\usepackage{pbox}
\usepackage{booktabs}
\usepackage{makecell} 
\usepackage{threeparttable} 
\usepackage{balance}
\usepackage[justification=centering,labelsep=period]{caption} 
\usepackage{color} 
\usepackage{gensymb} 
\usepackage{upgreek}
\hypersetup{colorlinks=true, linkcolor=black} 

\usepackage{adjustbox}
\usepackage{xcolor}
\usepackage{soul}
\soulregister\cite7 
\soulregister\citep7 
\soulregister\citet7 
\soulregister\ref7 
\soulregister\pageref7 

\usepackage{bm}
\usepackage{geometry}
\title{
    ECMD: An Event-Centric Multisensory Driving Dataset for SLAM
}

\author{Peiyu~Chen$^{1*}$, Weipeng~Guan$^{1*}$, Feng~Huang$^{2*}$, Yihan~Zhong$^{2}$, Weisong~Wen$^{2}$, Li-Ta~Hsu$^{2}$, Peng~Lu$^{1 \text{\textdagger}}$     
    \thanks{
        *Equal contribution; 
         $^{\text{\textdagger}}$Corresponding Author.
        
        $^{\text{1}}$Peiyu Chen, Weipeng Guan, and Peng Lu are with the Adaptive Robotic Controls Lab (ArcLab), Department of Mechanical Engineering, The University of Hong Kong, Hong Kong SAR, China. (e-mail: chenpyhk@connect.hku.hk; wpguan@connect.hku.hk; lupeng@hku.hk). 

        $^{\text{2}}$Feng Huang, Yihang Zhong, Weisong Wen, and Li-Ta Hsu are with the Intelligent Positioning and Navigation Laboratory, Department of Aeronautical and Aviation Engineering, The Hong Kong Polytechnic University, Hong Kong SAR, China. (e-mail: darren-f.huang@connect.polyu.hk; yihan1.zhong@connect.polyu.hk; welson.wen@polyu.edu.hk; lt.hsu@polyu.edu.hk)
        }%
    \thanks{
        This work was supported by General Research Fund under Grant 17204222, and in part by the Seed Fund for Collaborative Research and General Funding Scheme-HKU-TCL Joint Research Center for Artificial Intelligence.
        }%
}

\begin{document}   
\maketitle  
\thispagestyle{headings} 
\pagestyle{headings}

\begin{abstract}
Leveraging multiple sensors enhances complex environmental perception and increases resilience to varying luminance conditions and high-speed motion patterns, achieving precise localization and mapping.
This paper proposes, ECMD, an event-centric multisensory dataset containing 81 sequences and covering over 200 km of various challenging driving scenarios including high-speed motion, repetitive scenarios, dynamic objects, etc.
ECMD provides data from two sets of stereo event cameras with different resolutions (640$\times$480, 346$\times$260), stereo industrial cameras, an infrared camera, a top-installed mechanical LiDAR with two slanted LiDARs, two consumer-level GNSS receivers, and an onboard IMU.
Meanwhile, the ground-truth of the vehicle was obtained using a centimeter-level high-accuracy GNSS-RTK/INS navigation system.
All sensors are well-calibrated and temporally synchronized at the hardware level, with recording data simultaneously.
We additionally evaluate several state-of-the-art SLAM algorithms for benchmarking visual and LiDAR SLAM and identifying their limitations.
The dataset is available at \url{https://arclab-hku.github.io/ecmd/}.
%
\end{abstract}


\section{Introduction}
\label{Introduction}

\begin{figure}[htb]  
        \centering
        \captionsetup{justification=justified}
        \includegraphics[width=1.0\columnwidth]{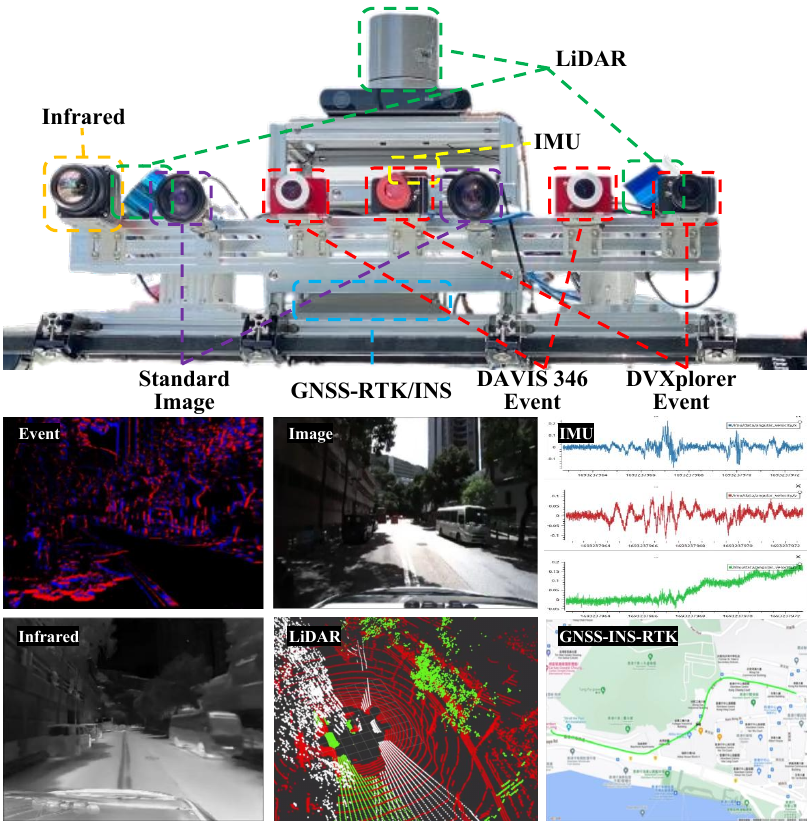}
        \caption{An overview of the sensor setup and dataset visualization. Above: The sensor suite is mounted on the top of the vehicle. Below: Sensor systems include two sets of stereo event cameras, stereo industrial cameras, an infrared camera, a top-installed LiDAR with two slanted LiDAR, IMU, and GNSS-RTK/INS systems. Each sensor is indicated with the letter box.}  
        \label{sensor_setup}
        \vspace{-2.0em}
\end{figure}%
\IEEEPARstart{V}{isual} and LiDAR simultaneous localization and mapping (SLAM) achieved notable progress within driving scenarios in recent years.
However, they encounter the challenging task of operating robustly under heterogeneous environments, such as varying lighting conditions, low-texture scenarios, repetitive structures, diverse motion patterns, dense dynamic objects, etc.
Utilizing novel sensors and integrating multiple sensors can provide a comprehensive perception and enhance the robustness of the entire system\cite{liu2023GLIO, MC-VEO, yan2023integration}.
These motivate us to develop a dataset that integrates novel sensors under realistic and complex driving scenarios, thereby promoting SLAM research.
 
Event cameras have low latency ($\upmu$s-level) and high dynamic range (140 dB compared
to 60 dB with standard cameras) properties, which offers great opportunities for visual (VO) and visual-inertial odometry (VIO) in rough terrain, aggressive motions, and high dynamic range (HDR)\cite{EVENT-SURVEY}.
Unlike traditional frame-based cameras that directly capture fixed-rate intensity frames, event cameras are motion-activated sensors that capture pixel-wise intensity differences asynchronously in continuous streams.
However, the widespread commercialization and implementation of event cameras in robotics are still early due to the expensive cost. 
In addition, event cameras confront challenges during rapid vibrations and ego-motion, as these conditions generate a substantial quantity of events, leading to intensive computations. 
Conversely, in cases where minimal relative motion between the event camera and the scene exists, such as under static states, they only provide limited information or even introduce noise\cite{PL-EVIO}.
Therefore, we embark on this research effort to explore the inquiry: Are event cameras ready for autonomous driving?

There exist several stereo event-based driving datasets that are worth mentioning and exploring.
MVSEC\cite{MVSEC} was the first stereo event-based driving dataset proposed for evaluating the localization performance.
While MVSEC employs the low resolution of DAVIS346 which limits the feature detection for accurate localization.
DSEC\cite{DSEC} offers stereo event streams with a high resolution of 0.31 Megapixels(MP). 
However, this dataset focuses on computer vision tasks segmentation, depth estimation, optical flow estimation, etc., which is not specifically designed for VO/VIO/SLAM domains.
MA-VIED\cite{MA-VIED2023} propose a large-scale driving dataset under standard urban scenarios and race track-like loops.
The ground-truth trajectory relies on GNSS-RTK, which only ensures high accuracy in open-sky environments and fails to provide high accuracy in GNSS-denied scenarios such as tunnels or densely street areas.
Ref.\cite{hadviger2023stereo} focuses on collecting both stereo event data and stereo intensity images under indoor and urban driving scenes with the ground-truth of GNSS-RTK/INS. 
Their sequences do not encompass extremely high-speed or repetitive scenarios that could be challenging to VO/VIO/SLAM algorithms. 


%

To address the above drawbacks, we propose ECMD, a dataset procured from diverse challenging driving scenarios with a comprehensive suite of sensors for benchmarking various VO/VIO/SLAM algorithms.
To the best of our knowledge, this is the first event-based SLAM dataset specifically focused on densely urbanized driving scenarios.
The contributions of our work can be summarized as follows:
\begin{enumerate}

\item 
Our sensor platform consists of various novel sensors shown in Fig.\ref{sensor_setup}, including two sets of stereo event cameras with distinct resolutions (640$\times$480, 346$\times$260), 
an infrared camera, 
stereo industrial cameras, 
three mechanical LiDARs (including two slanted LiDARs), 
a high-quality inertial measurement unit (IMU), 
and three global navigation satellite system (GNSS) receivers.
For the ground-truth, we adopt a centimeter-level position system that combines the GNSS real-time kinematic (RTK) with the fiber optics gyroscope integrated inertial system as GNSS-RTK/INS. 

\item 
ECMD collects 81 sequences covering over 200 kilometers of trajectories in various driving scenarios, including dense streets, urban, tunnels, highways, bridges, and suburbs. 
These sequences are recorded under daylight and nighttime, providing challenging situations for Visual and LiDAR SLAM, e.g., dynamic objects, high-speed motion, repetitive scenarios, and HDR scenes.
Meanwhile, we evaluate existing state-of-the-art visual and LiDAR SLAM algorithms with various sensor modalities on our datasets.
Moreover, our dataset and benchmark results are released publicly available on our website.


\end{enumerate}

The remainder of the paper is organized as follows:
Section \ref{Related works} introduces the related works. 
Section \ref{System Overview} presents the sensor setup and sensor calibration.
Section \ref{Dataset Overview} introduces the dataset overview.
Section \ref{Dataset Applications} demonstrates the dataset application.
Section \ref{ISSUES} introduces known issues.
The conclusion is given in Section \ref{CONCLUSIONS}.

\section{Related works}
\label{Related works}
Currently, several event-based datasets combined with various sensors have been released for VO/VIO/SLAM domains, utilizing handheld devices or a variety of robotics platforms.
DAVIS240C\cite{DAVIS240c}, TUM-VIE\cite{TUM-VIE}, VECtor\cite{VECtor}, and HKU-dataset\footnote{\url{https://github.com/arclab-hku/Event_based_VO-VIO-SLAM}} were collected by handheld / head-mounted devices under indoor environments.
M2DGR\cite{M2DGR} utilizes ground robots to collect a multi-sensor dataset with an event camera under large-scale scenes, while the event streams exhibit large noises.
FusionPortable\cite{FusionPortable} proposes multi-sensor campus-scene datasets with stereo event cameras on diverse platforms (handheld, quadruped robot, and UGV).
Moreover, there exist specialized event-based datasets such as UZH-FPV\cite{UZH-FPV} and GRIFFIN\cite{GRIFFIN}, which are targeted for flying robots.

Moreover, a number of event-based datasets are published under large-scale driving scenarios for computer vision.
These autopilot datasets offer more realistic and challenging conditions, including high-speed scenarios, repetitive situations, and HDR scenes compared to datasets collected from handheld devices.
The first dataset catering to driving recordings using an event camera is DDD17\cite{DDD17}, as well as the follow-up DDD20\cite{DDD20}, for studying the end-to-end driving application incorporating diverse vehicle control data.
HATS\cite{HATS}, CED\cite{CED}, Ref.\cite{rebecq2019high}, and Ref.\cite{Brisbane-Event-VPR} published their event-based datasets for the computer vision task of object classification, image reconstruction, and vision place recognition in driving scenarios.
MVSEC\cite{MVSEC} is a pioneering cross-modal dataset with stereo event and image cameras, as well as LiDAR. 
However, a limitation of MVSEC resides in the utilization of low-resolution event cameras (346$\times$260) with a compact baseline of 10 cm, coupled with the imprecision of the ground-truth derived from GNSS or LiDAR-SLAM.
DSEC\cite{DSEC} proposed an event-based dataset whose scenarios are similar to KITTI\cite{KITTI}, providing higher resolution stereo event (640$\times$480) and image, LiDAR, and IMU under various illumination conditions.
M3ED\cite{M3ED} encompasses high-resolution event cameras (1280$\times$720) and covers three different robotics platforms: driving, flight, and legged robot. 
However, both DSEC and M3ED datasets are primarily utilized for computer vision fields, such as optical flow estimation, segmentation, and disparity estimation, rather than specifically for localization or mapping problems.
Besides, they do not provide sufficient challenges for SLAM, as the majority of these datasets were collected in rural or suburban areas with relatively low-lying structures, light traffic, and less dynamic objects.
ViViD++\cite{ViViD++} focuses on diverse vision sensors for handheld and driving platforms, including event, thermal, and standard cameras.
MA-VIED\cite{MA-VIED2023} proposes a comprehensive driving dataset that encompasses race track-like loops, maneuvers, and standard driving scenarios. 
However, both of these datasets exclusively offer monocular data for each camera type, thereby precluding the possibility of conducting stereo visual SLAM.
Ref.\cite{hadviger2023stereo} introduces a stereo visual localization dataset that exploits both the high-resolution event and standard cameras under indoor and urban scenarios.

Table \ref{literature_comparison}. summarizes the differences between our ECMD and other event-based datasets under autonomous driving scenarios.
Compared to other datasets, our ECMD offers several advantages: 
(\romannumeral1) Capture diverse visual data format (RGB image, event stream, and infrared image) from multiple types of vision sensors in varying luminance conditions and urbanized scenarios;
(\romannumeral2) \textsl{1kHz-rate} event streams from different resolution event cameras empower in-depth exploration of event-based perception;
(\romannumeral3) Based on our previous work\cite{UrbanNav}, three LiDARs, including two slanted LiDARs, are employed to collect high-rising building structures for LiDAR point cloud maps generation;
(\romannumeral4) We employ a centimeter-level localization system, GNSS-RTK/INS, as ground-truth, enabling a comprehensive evaluation of various SLAM algorithms.

\begin{table*}[htbp]
        \renewcommand\arraystretch{1.2}
        \begin{center}
        \caption{Comparison with other event-based datasets in driving scenarios, regarding terrain and sensor types.
        ECMD provides the most extensive sensor configuration and urban scenarios.
        }
        \label{literature_comparison}
        \resizebox{\columnwidth*2}{!}
        { 
        \begin{threeparttable}
        \begin{tabular}{ccccccccc} 
        \hline\hline  
        \multicolumn{1}{c}{\multirow{2}*{\textbf{Dataset}}}
    & \multicolumn{1}{c}{\multirow{2}*{\textbf{Terrain}}}
    & \multicolumn{1}{c}{\multirow{1}*{\textbf{Event}}}
    & \multicolumn{1}{c}{\multirow{1}*{\textbf{Infrared}}}
    & \multicolumn{1}{c}{\multirow{1}*{\textbf{Image}}}
    & \multicolumn{1}{c}{\multirow{2}*{\textbf{LiDAR}}}
    & \multicolumn{1}{c}{\multirow{2}*{\textbf{GNSS}}}
    & \multicolumn{1}{c}{\multirow{2}*{\textbf{IMU}}}
    & \multicolumn{1}{c}{\multirow{2}*{\textbf{GT Pose}}}\\
    \cline{3-5}
    & &  \multicolumn{3}{c}{\multirow{1}*{Resolution [MP]}}  &  & & &
    \\   
\hline
DDD17\cite{DDD17}       & Urban         & 0.09 $\times$ 1   & \ding{55}    & 0.09 $\times$ 1 & \ding{55}    & \ding{51}    & \ding{55} & \ding{55}   \\
\hline
N-CARS\cite{HATS}       & Urban         & 0.07 $\times$ 1  & \ding{55}     & 0.07 $\times$ 1 & \ding{55}    & \ding{55}    & \ding{55} & \ding{55}   \\
\hline
MVSEC\cite{MVSEC}       & Suburban      & 0.09 $\times$ 2  & \ding{55}     & 0.36 $\times$ 2 & VLP-16       & \ding{51}    & \ding{51} & GNSS/LiDAR-SLAM   \\
\hline
CED\cite{CED}           & Urban         & 0.09 $\times$ 1  & \ding{55}     & 0.09 $\times$ 1 & \ding{55}    & \ding{55}    & \ding{55} & \ding{55}   \\
\hline
Ref.\cite{rebecq2019high}& Urban         & 0.31 $\times$ 1  & \ding{55}     & \ding{55} & \ding{55}    & \ding{55}    & \ding{55} & \ding{55}   \\
\hline
DDD20\cite{DDD20}       & Urban         & 0.09 $\times$ 1  & \ding{55}     & 0.09 $\times$ 1 & \ding{55}    & \ding{51}    & \ding{55} & \ding{55}   \\
\hline
Brisbane-Event-VPR\cite{Brisbane-Event-VPR} & Suburban      & 0.31 $\times$ 1  & \ding{55}     & 0.31 $\times$ 1 & \ding{55} & \ding{55} & \ding{55} & \ding{55}   \\
\hline
Ref.\cite{de2020large} & Suburban, Urban      & 0.31 $\times$ 1  & \ding{55}     & 0.31 $\times$ 1 & \ding{55} & \ding{55} & \ding{55} & \ding{55}   \\
\hline
DSEC\cite{DSEC}         & Suburban      & 0.31 $\times$ 2   & \ding{55}     & 1.56 $\times$ 2 & VLP-16 & \ding{51}    & \ding{51} & GNSS-RTK     \\
\hline
ViViD++\cite{ViViD++}   & Urban         &  0.31 $\times$ 1 & 0.33 $\times$ 1  & 2.46 $\times$ 1 &  OS1-64 & \ding{51} & \ding{51} & GNSS-RTK   \\
\hline
M3ED\cite{M3ED}         & Forest, Urban & 0.92 $\times$ 2 & \ding{55}     & 1.02 $\times$ 3 & OS1-64 & \ding{55}    & \ding{51} & LiDAR-SLAM/GNSS-RTK  \\
\hline
MA-VIED\cite{MA-VIED2023}         &  \makecell{Urban, \\ Race track-like loops} & 0.31 $\times$ 1 & \ding{55}     & 2.30 $\times$ 1 & \ding{55} & \ding{51}    & \ding{51} & GNSS-RTK  \\
\hline
Ref. \cite{hadviger2023stereo}         & Urban, Indoor  & 0.31 $\times$ 2 & \ding{55}     & 3.14 $\times$ 2 & OS1-128 & \ding{51}    & \ding{51} & GNSS-RTK/INS  \\
\hline
\textbf{ECMD}   & \makecell{Suburban, Urban, \\ Dense City} & \makecell{0.09 $\times$ 2 \\ 0.31 $\times$ 2}  & 0.33 $\times$ 1  & 2.30 $\times$ 2 & \makecell{VLP-16 \\ Lslidar C16\\ HDL-32E} & \ding{51} & \ding{51} & GNSS-RTK/INS   \\
\hline\hline        
        \end{tabular}
        \end{threeparttable} 
        }
        \end{center}
\end{table*}

\section{System Overview}
\label{System Overview}


\subsection{Sensors Setup}
The data collection platform is shown in Fig.\ref{sensor_setup}. Our sensor suite consists of a multi-camera setup (event camera, industrial camera, and infrared camera) equipped with three LiDARs, high-quality IMU, three GNSS receivers, and GNSS-RTK/INS systems.
The specific specifications of each sensor are presented in Table \ref{sensor_param}.
An Intel NUC (i7-1260P, 32GB RAM) and an industrial computer (i7-10610U, 32GB RAM) are used to run sensor drivers, and record data into ROS bags on the Ubuntu system.

\subsubsection{Visual Sensors}
Two sets of stereo event cameras with different resolutions, DAVIS436 (346$\times$260) and DVXplorer (640$\times$480), are configured at a baseline of 30 cm respectively.
DAVIS346 produces asynchronous events and intensity frames. 
In contrast, DVXplorer exclusively generates events, while its resolution surpasses that of DAVIS346, enabling the provision of more intricate scene information.
Each event camera is equipped with additional infrared filters to mitigate interference from LiDAR.
Two FLIR BFLY-U3-23S3C industrial cameras with a resolution of 1920$\times$1200 are used to capture RGB images at 20 Hz in fixed exposure mode.
Forward-facing stereo industrial cameras are installed with a baseline of 30 cm, ensuring fairness by maintaining consistency with the baseline of the stereo event cameras.
Hikrobot MV-CI003-GL-N6 infrared camera collects thermal frames at 20 Hz, encompassing a response band of 8-14$\upmu$m and equipped with a 6.3mm focal length lens.

\subsubsection{Mechanical LiDAR}
We configure three mechanical LiDARs including two slanted LiDARs to collect accurate point clouds of surrounding environments.
Velodyne HDL-32E is positioned on the top of the vehicle to capture the surroundings horizontally.
Two slanted LiDARs, Lslidar C16 and Velodyne VLP-16, are mounted on the left and right sides of the sensor kit, respectively.
This configuration facilitates the thorough recording of architectural particulars relevant to high-rising buildings in urbanized areas and all LiDAR data are collected at 10 Hz.

\subsubsection{GNSS-RTK/INS Sensor}
A tactical-level Xsens-MTI-30 IMU is employed to collect the raw acceleration and angular velocity at 400 Hz.
The accurate ground-truth of localization is furnished by a centimeter-level GNSS-RTK/INS navigation system, further details can be found in Section \ref{Ground-truth Poses}.

\begin{table}[htbp]
        \renewcommand\arraystretch{1.2}
        \begin{center}
        \caption{Sensors specifications}
        \label{sensor_param}
        \resizebox{\columnwidth*1}{!}
        { 
        \begin{threeparttable}
        \begin{tabular}{cl} 
        \hline\hline  
        Sensors & Specifications \\
        \hline
        \multirow{8}{*}{Event Camera} 
         & Inivation DAVIS346 Color ($\times$2), 1000 Hz \\
         & 346$\times$260 pixel \\
         & IMU: MPU6150, 6-axis \\ 
         & baseline: 30 cm\\
        \cline{2-2}
         & Inivation DVXplorer ($\times$2), 1000 Hz \\
         & 640$\times$480 pixel \\
         & IMU: MPU6150, 6-axis \\ 
         & baseline: 30 cm\\
         \hline
        \multirow{3}{*}{Infrared Camera}
         & Hikrobot MV-CI003-GL-N6, 20 Hz \\
         & 640$\times$512 pixel \\
         & H-FOV: 88.5$^\circ$, V-FOV: 73.2$^\circ$ \\ 
         \hline
         \multirow{4}{*}{Industrial Camera} &  FLIR BFLY-U3-23S3C($\times$ 2), 20 Hz\\
         & 1920$\times$1200 pixel \\
         & H-FOV: 96.8$^\circ$, V-FOV: 79.4$^\circ$ \\ 
         & baseline: 30 cm\\
         \hline
        \multirow{12}{*}{LiDAR} 
         & Velodyne HDL-32E, 10Hz \\
         & H-FOV: 360$^\circ$, V-FOV: +10.67$^\circ$$\sim$-30.67$^\circ$ \\
         & 32 channel \\
         & 100m range\\ \cline{2-2} 
         & Velodyne VLP-16, 10Hz \\
         & H-FOV: 360$^\circ$, V-FOV: +15$^\circ$$\sim$-15$^\circ$ \\
         & 16 channel \\
         & 100m range\\ \cline{2-2} 
         & Lslidar C16, 10Hz \\
         & H-FOV: 360$^\circ$, V-FOV: +14$^\circ$$\sim$-16$^\circ$ \\
         & 16 channel \\
         & 150m range\\ 
         \hline
        \multirow{3}{*}{IMU} & Xsens Mti-30, 400Hz\\
         & Accelerometer in-run Bias Instability 15 \si{\micro\gram} \\
         & Gyroscope in-run Bias Instability 18$^\circ$/h \\ \hline
        \multirow{2}{*}{GNSS Receiver} & U-Blox ZED-F9P, 1Hz \\
         & EVK-M8T, 1Hz \\
         \hline
        \multirow{2}{*}{GNSS-RTK/INS} & NovAtel SPAN-CPT \cite{NovAtel}, 1Hz \\
         & Localization RMSE 5cm \\
        \hline\hline        
        \end{tabular}
        \end{threeparttable} 
        }
        \end{center}
\end{table}

\subsection{Time Synchronization}

We use a Precision Time Protocol (PTP)\cite{PTP} device to synchronize the clocks of various data collection devices across the sensor network. The PTP ensures time accuracy within nanoseconds. 
The synchronization device acquires the NMEA\cite{langley1995nmea} output and pulse-per-second (PPS) signal from a u-blox M8T GNSS receiver to align the ROS time of the onboard computers with the GPS time. This enables sensors such as cameras, LiDAR, and IMU to record timestamps based on the synchronized GPS time. 
Moreover, to achieve time synchronization between different event cameras, the DAVIS346 on the rightmost side is configured as the master device and transmits trigger signal pulses to the remaining slave event cameras sequentially from left to right via external cables.

\subsection{Sensors Calibration}

\subsubsection{IMU Calibration}
To calibrate the IMU, we position it on a level surface for a duration of three hours and record the raw measurements. Utilizing the Kalibr toolbox, we can accurately calibrate the random walk and Gaussian white noise of the IMU.

\subsubsection{Industrial Cameras Calibration}
For industrial cameras, we move the sensor platform against the 9$\times$7 checkerboard in the XYZ-axis and collect the sequence of RGB images and IMU.
Then intrinsics calibration of industrial cameras is achieved by Kalibr toolbox\cite{Kalibr2013}, where the pinhole and radial-tangential camera models are adopted. 

\subsubsection{Event Cameras Calibration}
\begin{figure}[htb]  
    	\subfigtopskip=0pt 
    	\subfigbottomskip=0pt 
    	\subfigcapskip=-8pt 
        \captionsetup{justification=justified}
        \centering
        \subfigure[]{
                \begin{minipage}[t]{0.48\columnwidth}
                \centering
                \includegraphics[width=1.0\columnwidth]{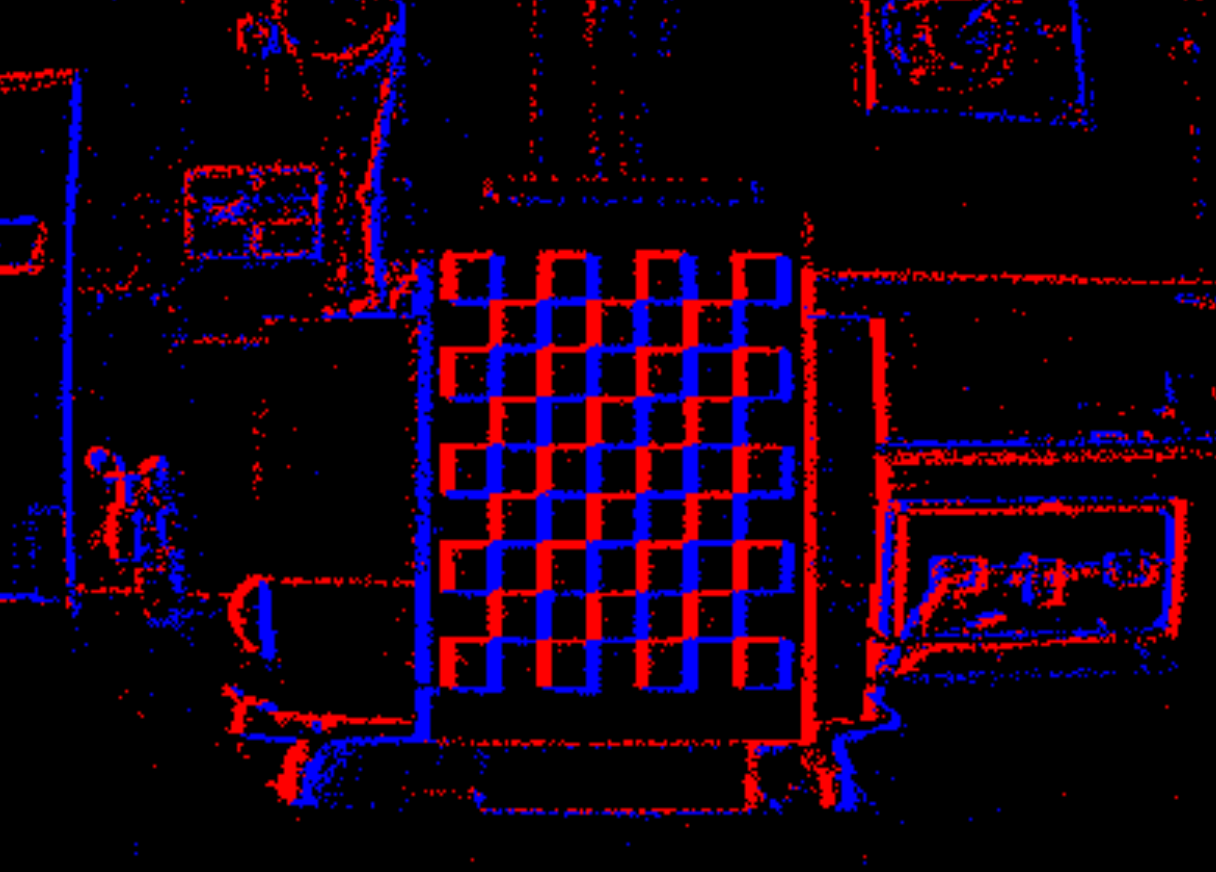}
                \label{event_stream}
                \end{minipage}%
        }%
        \subfigure[]{
                \begin{minipage}[t]{0.48\columnwidth}
                \centering
                \includegraphics[width=1.0\columnwidth]{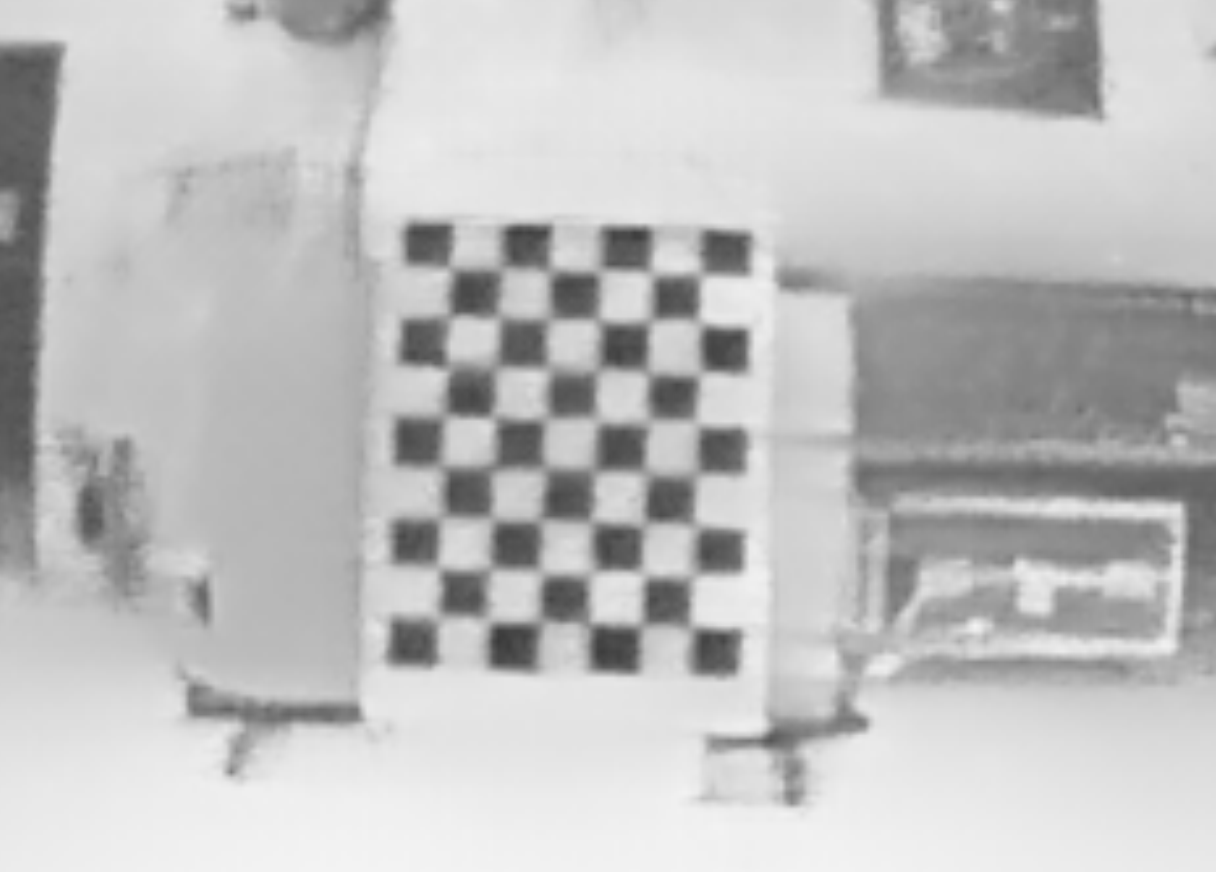}
                \label{event_reconstruction}
                \end{minipage}%
        }%
        \caption{The calibration of event cameras using checkerboard. (a) Raw event streams. (b) Reconstructed images from events.}
        \label{event_calibration}
\end{figure}

For event cameras, DAVIS346 can produce fixed-rated frames, enabling image-based calibration, while DVXplorer merely produces asynchronous event streams.
Therefore, E2Calib\cite{event2021calibrate}\cite{eventreconstruct} is used to achieve image reconstruction from event streams.
With the reconstructed checkerboard images in Fig. \ref{event_reconstruction}, the intrinsics of event cameras could also be calibrated by Kalibr.

\subsubsection{Infrared Camera Calibration}
\begin{figure}[htb]  
    	\subfigtopskip=0pt 
    	\subfigbottomskip=0pt 
    	\subfigcapskip=-8pt 
        \captionsetup{justification=justified}
        \centering
        \subfigure[]{
                \begin{minipage}[t]{0.48\columnwidth}
                \centering
                \includegraphics[width=1.0\columnwidth]{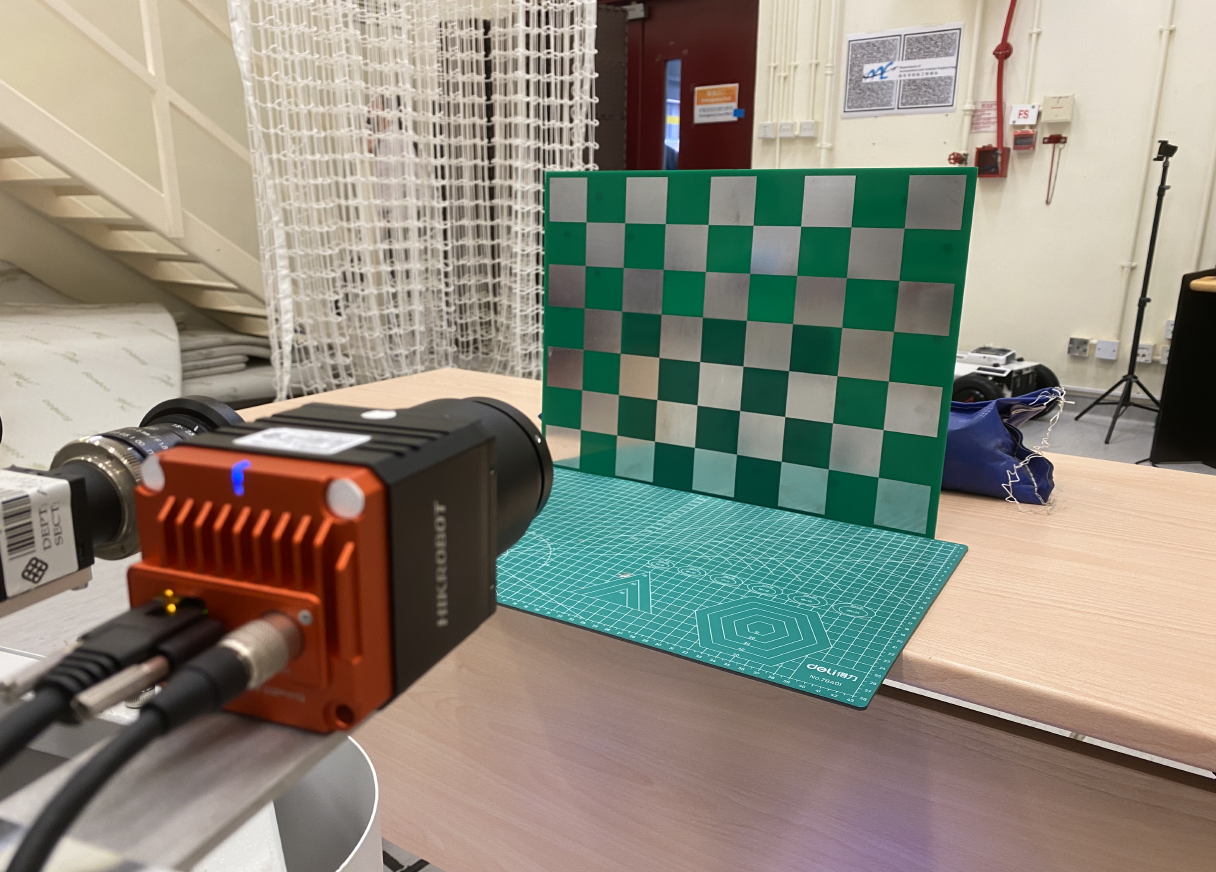}
                \label{pcb_board}
                \end{minipage}%
        }%
        \subfigure[]{
                \begin{minipage}[t]{0.48\columnwidth}
                \centering
                \includegraphics[width=1.0\columnwidth]{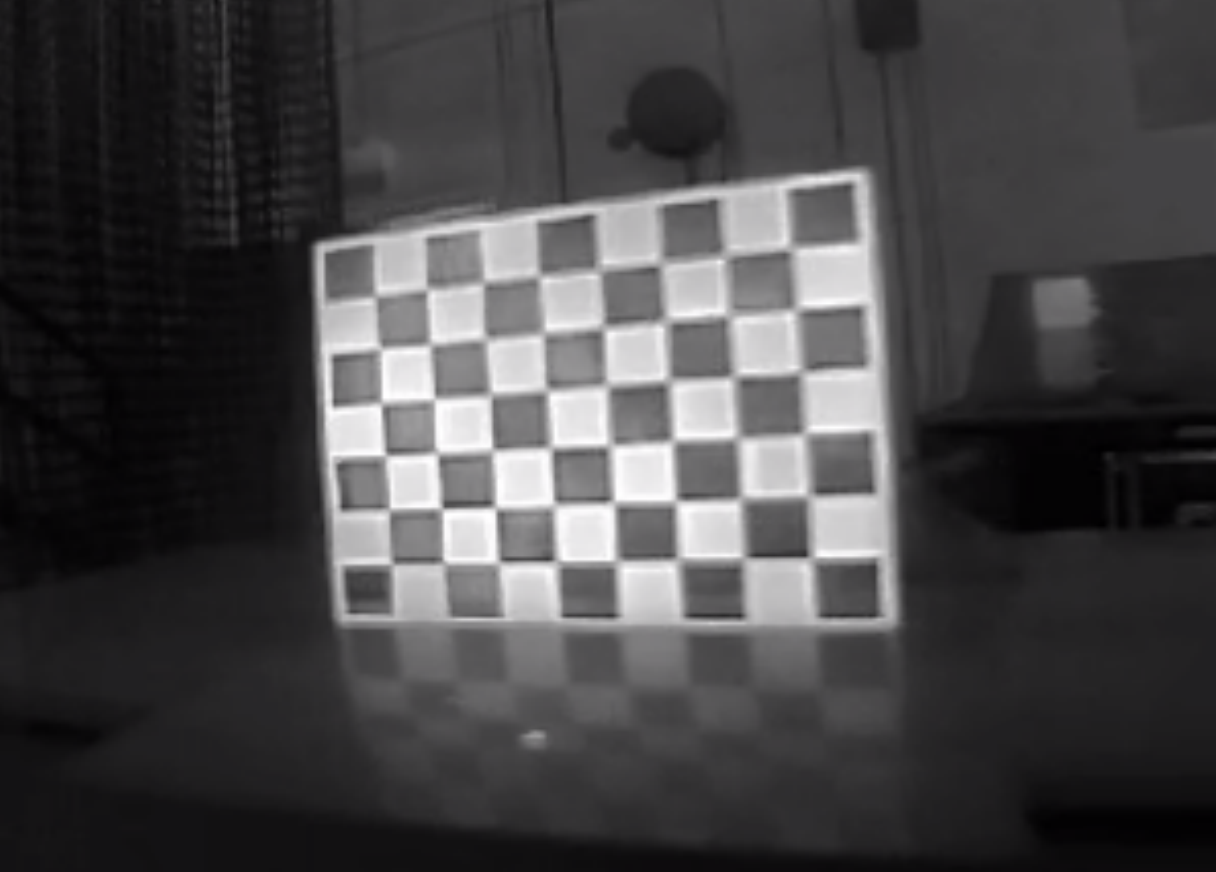}
                \label{infrared_image}
                \end{minipage}%
        }%
        \caption{The calibration of the infrared camera using checkerboard. (a) PCB checkerboard. (b) Infrared image.}
        \label{infrared_calibration}
\end{figure}

Due to infrared cameras solely capturing the temperature rather than the intensity difference, we design a distinct 9$\times$7 checkerboard to make the pattern detectable for infrared cameras. 
As shown in Fig.\ref{pcb_board}, the checkerboard intervals are affixed with aluminum materials, and then using a heating plate to raise the temperature of the checkerboard.
Since the superior thermal dissipation of aluminum compared to plastic, a temperature contrast emerges between the two materials, enabling infrared cameras to distinctly capture the lattice shape of the checkerboard, as in Fig.\ref{infrared_image}.
With the special infrared image of the checkerboard, intrinsic can be calibrated by Kalibr.

\subsubsection{Calibration between Camera and IMU}
After completing intrinsics calibration, we move the sensor suite in front of checkerboards along the XYZ-RPY-axis and collect data simultaneously.
Subsequently, the extrinsics and the temporal offset between all cameras and IMU could be estimated using Kalibr.

\subsubsection{Calibration between LiDAR and IMU}
\begin{figure}[htb]  
        \centering
        \captionsetup{justification=justified}
        \includegraphics[width=0.8\columnwidth]{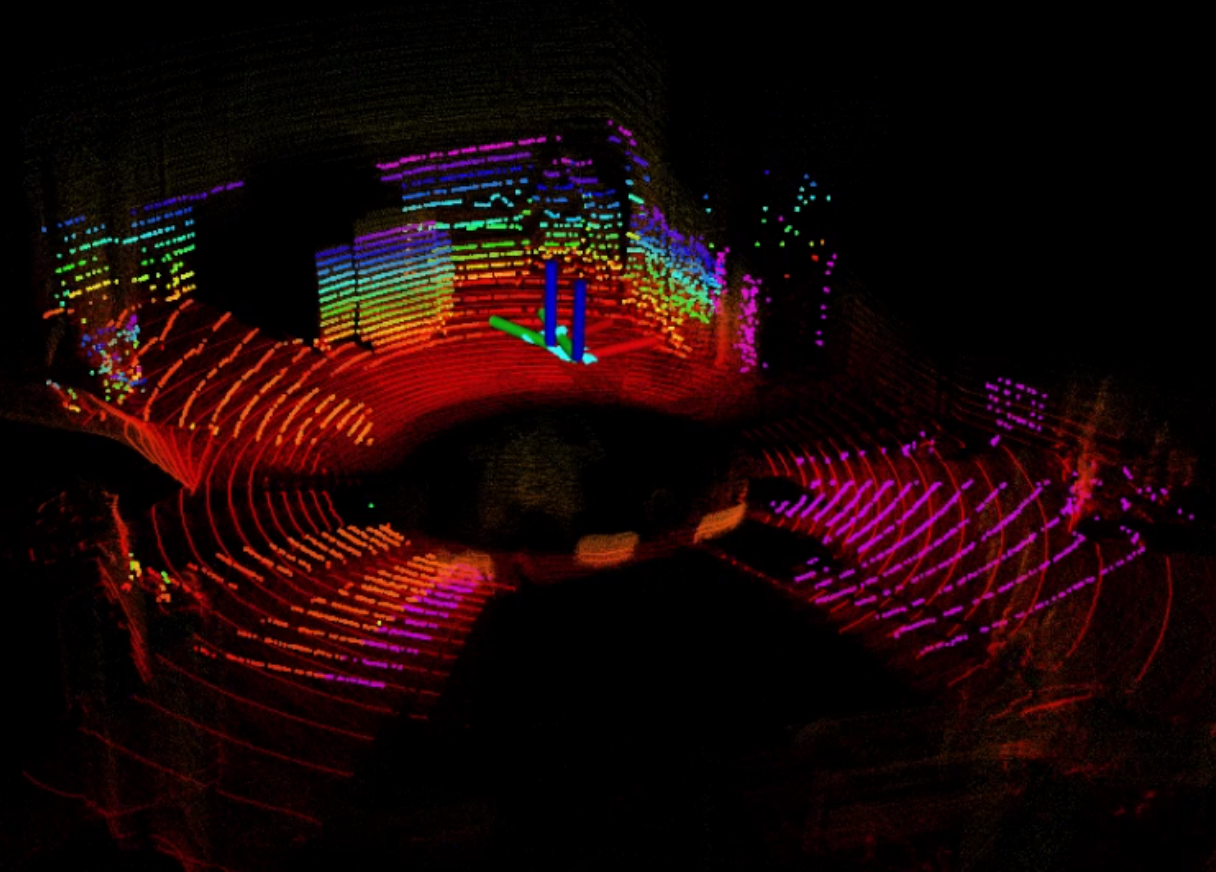}
        \caption{Calibration between LiDAR and
IMU.}  
        \label{lidar_calibration}
        \vspace{-2.0em}
\end{figure}

For the calibration of mechanical LiDAR, LI-Init \cite{LiDARCalibrMars2022} is capable of achieving temporal and spatial calibration for LiDAR and IMU without checkerboards or extra devices in Fig.\ref{lidar_calibration}. 
We rotate and move the device around the XYZ-axis to ensure sufficient excitation until the data accumulation is completed, thus we acquire the extrinsic transformation between LiDAR and IMU.

\section{Dataset Overview}
\label{Dataset Overview}
Our dataset encompasses a wide range of driving scenes, including urban streets, urban roads, tunnels, highways, bridges, and suburban roads.
We have specifically focused on scenarios where visual SLAM algorithms encounter difficulties. 
These scenarios involve high-speed motion (up to 110 km/h), limited texture, as well as difficult glare conditions in both daytime and nighttime driving.
We also targeted situations where LiDAR SLAM encounters limitations, such as long corridors or areas with sparse geometric structures.
The complete dataset is partitioned into 81 sequences to facilitate researchers in evaluating their algorithms. 
Each sequence has an approximate duration of 120 seconds. 
Additionally, we have retained a few sequences with long duration, lasting approximately 34 minutes, specifically for the evaluation of loop closure in large-scale environments and loop closure scenarios.
The summary of sequence types can be found in Table \ref{dataset sequence}.

\begin{table}[htbp]
        \setlength{\abovecaptionskip}{-0.02cm}
        \renewcommand\arraystretch{1.2}
        \tiny 
        \begin{center}
        \caption{The summary of different sequence types under contrasting lighting conditions}
        \label{dataset sequence}
        \resizebox{\columnwidth*1}{!}
        { 
        \begin{threeparttable}
        \begin{tabular}{cccc} 
        \hline  
        Terrain & Time & Duration [min]& \#Sequences\\
        \hline
        Dense Street & Day, Night& 91 &  19 \\
        Urban Road & Day, Night & 35.5 & 19 \\
        Tunnel & - - - & 6.5 & 9 \\
        Highway & Day, Night & 34.3& 15 \\
        Bridge & Day, Night & 10.5 & 6 \\
        Suburban Road & Day, Night & 23.5 & 13 \\
        \hline
        Total & Day, Night& 201.3 & 81\\
        \hline
        \end{tabular}
        \end{threeparttable} 
        }
        \end{center}
        \vspace{-1.0cm}
\end{table}

\subsection{Scenarios}
\begin{figure*}[htb]  
        \centering
        \captionsetup{justification=justified}
        \includegraphics[width=2.0\columnwidth]{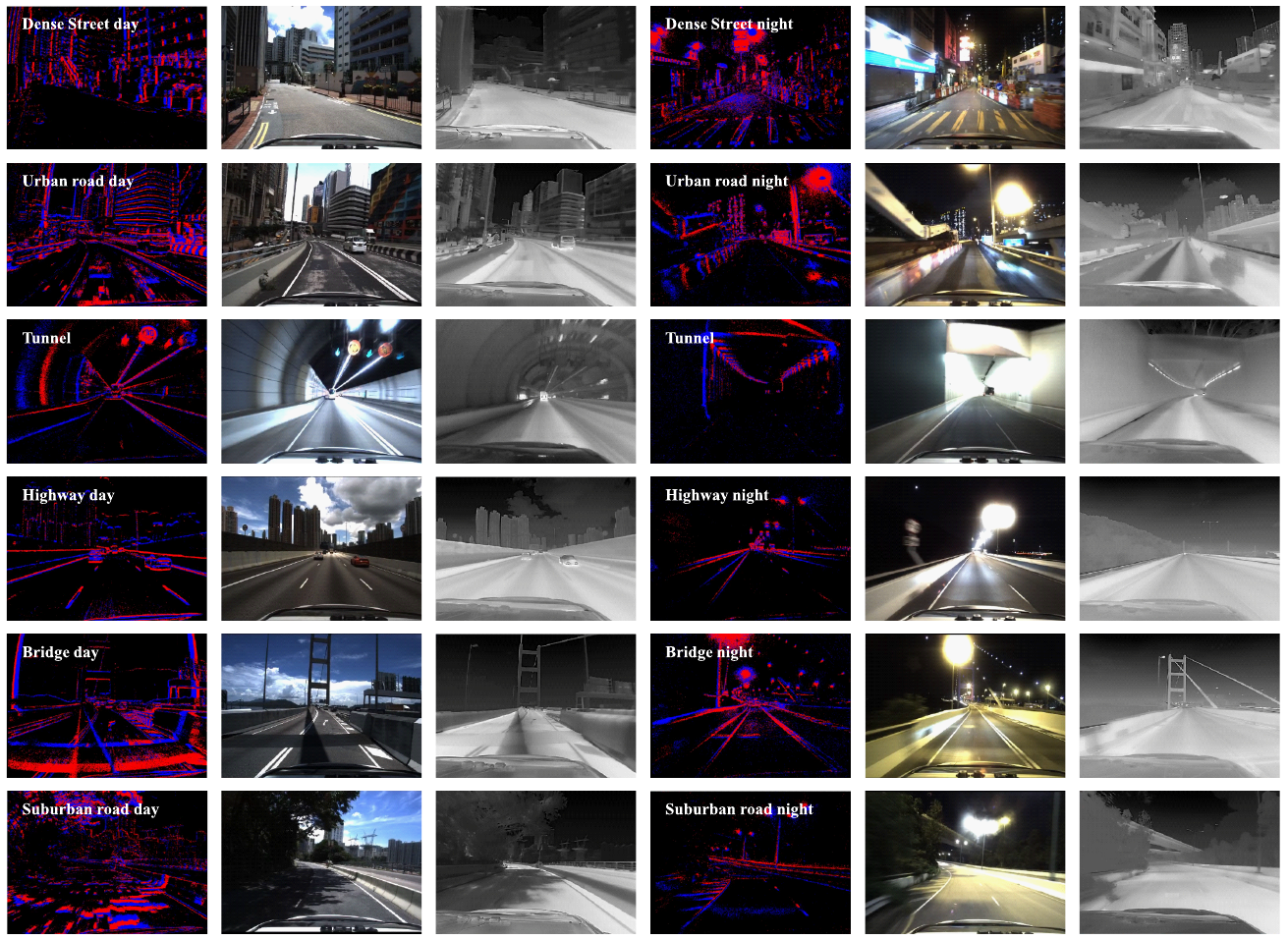}
        \caption{The visualization of various scenarios including event streams, RGB images, and infrared images. }  
        \label{Scene}
        \vspace{-1.0em}
\end{figure*}%

\subsubsection{Dense\_Urban\_Street}

This scenario focuses on low-speed vehicles, around 30km/h, proceeding on highly urbanized areas and urban canyons in Hong Kong with multiple light conditions.
The streets are narrow at 10m in width and buildings on both sides of the scene are dense. 
Meanwhile, the presence of congested traffic and dynamic crowds may produce the degradation of visual or LiDAR localization, such as \textit{Dense\_street\_day\_easy\_b}.
To evaluate the loop closure performance of SLAM, we remarkably recorded sequences \textit{Dense\_street\_difficult\_circle} and \textit{Dense\_street\_difficult\_loop} where our vehicle was circling in repeated routes.

\subsubsection{Urban\_Road}
  
This type of scenario records the vehicle traveling at an approximate speed of 60km/h on an expressway in Hong Kong with multiple weather conditions.
Compared to the \textit{Dense\_Urban\_Street} scenario, Urban\_Road sequences travel through Hong Kong city at a higher speed, while the buildings are not as tightly packed on either side and the road is more spacious with four lanes.
Despite the absence of pedestrians on the road, the scene still includes vehicles overtaking, paralleling, and other situations where the relative motion is not consistent with the absolute motion.
The aforementioned discrepancy might pose a challenge for the VIO or LIO system.
Moreover, the sequence comprises the vehicle traveling during nighttime in rainy conditions.
We record trajectories in rainy situations under nighttime like \textit{Urban\_road\_night\_difficult\_rainy\_a} which are commonly faced in practical driving scenarios, whereas they are not present in previous datasets.

\subsubsection{Tunnel}
Tunnel scenarios commence with a high-speed vehicle on an open-sky highway, entering an enclosed tunnel without satellite reception.
Inside the tunnel, GNSS positioning is unreliable since the satellite signal is completely blocked.
Meanwhile, the scenario represents a typical and challenging scene for VIO and LIO systems due to the repetitive and texture-less environments for vision sensors and LiDAR.
The sequence collections end after the vehicle exits the tunnel and continues to proceed on the highway for twenty seconds.



\subsubsection{Highway}

The scenario involves vehicles traveling at speeds up to 100km/h on low-texture highways both during the day and night, with sparse buildings alongside the road.
High speeds, rapid changes in vehicle speed, repetitive visual scenes, and low-texture environments present significant challenges for autonomous driving. 
Meanwhile, the vibration of the vehicle body at high-speed motion amplifies the random walk and Gaussian white noise of IMU, thereby diminishing its reliability.

\subsubsection{Bridge}

The motion pattern of vehicles in bridge scenarios resembles that of highways, with vehicles traveling in a straight line at high speed along the bridge.
However, this scene differs as there are no buildings on either side of the bridge, only the sea surrounds it. 
Bridges present scenes with limited texture, and the feature information within these scenes tends to be monotonous and repetitive, which further exacerbates the challenge of achieving accurate localization.

\subsubsection{Suburban\_Road}

Suburban road scenarios present complex natural environments characterized by winding and rugged roads, steep slopes, and narrow lanes.
The vehicle navigates the serpentine mountain roads at a moderate speed (approximately 50km/h), with significant altitude changes.
The abundant texture information in the mountain road scene facilitates visual algorithms to extract stable features and construct effective constraints. 


\subsection{Ground-truth Generation}
\label{Ground-truth Generation}
\begin{figure*}[htb]  
        \centering
        \captionsetup{justification=justified}
        \includegraphics[width=2.0\columnwidth]{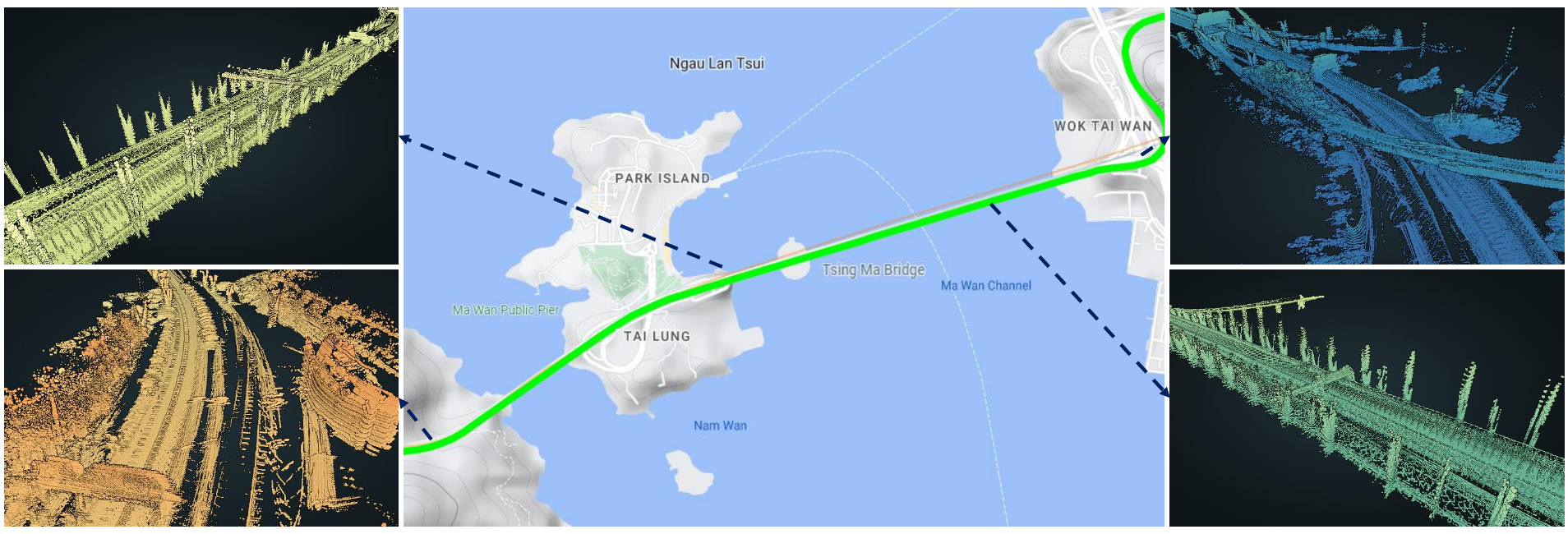}
        \caption{The vehicle poses ground-truth on Google map with the LiDAR point cloud maps of Tsing Ma bridge. }  
        \label{ground truth and maps}
\end{figure*}%

\subsubsection{Ground-truth Poses}
\label{Ground-truth Poses}

We obtained the ground-truth positioning from the NovAtel SPAN-CPT\cite{NovAtel}, a high-performance GNSS RTK/INS integrated navigation system.
The ground-truth of most existing event-based driving datasets are derived from LiDAR-SLAM\cite{MVSEC}\cite{M3ED}, GPS/GNSS\cite{MVSEC}, GNSS-RTK\cite{DSEC}\cite{ViViD++}\cite{M3ED}.
The ground-truth derived from LiDAR-SLAM relies on the estimation of vehicle trajectories using LiDAR SLAM which only provides relative trajectories.
It is difficult to quantify the accuracy of ground-truth pose, and errors may even exceed ten meters in some cases.
The complex environment or the equipment malfunctions may disrupt the satellite reception of GPS/GNSS, thus relying solely on GPS/GNSS for ground-truth pose may lead to significant drift.
The GNSS-RTK device can only provide centimeter-level accuracy in the open sky\cite{UrbanNav}
In contrast, our SPAN-CPT can provide continuous high accuracy aided by the internal fiber-optic gyroscopes under high-rise buildings, tunnels, and other environments with weak satellite signals. 
Furthermore, we post-process the ground-truth positioning from SPAN-CPT using the state-of-the-art NovAtel Inertial Explorer\cite{NovAtel} software to maximize the accuracy of the trajectory. 
For the GNSS positioning benchmark, we provide the WGS84 coordinate data for comparison. 
For the evaluation of SLAM algorithms, we provide the tools\footnote{\url{https://github.com/IPNL-POLYU/UrbanNavDataset/tree/master/tools/gt_vis}} to transform the ground-truth data from the WGS84 coordinates to the local frame/ENU frame based on the original points.

\subsubsection{LiDAR Point Cloud Maps Generation}
\label{LiDAR Point Cloud Maps Generation}
Utilizing the ground-truth pose for each frame in conjunction with their corresponding LiDAR point clouds, we accumulate these point clouds to construct a highly accurate LiDAR point cloud map to depict the TsingMa Bridge in Fig.\ref{ground truth and maps}. 
The map encompasses rich spatial information, providing a detailed 3D reconstruction of the bridge and its surrounding areas.

\section{Dataset Applications}
\label{Dataset Applications}

\subsection{Visual SLAM Evaluations}
\label{Visual SLAM Evaluations}
\begin{figure}[htb]  
        \centering
        \captionsetup{justification=justified}
        \includegraphics[width=1.0\columnwidth]{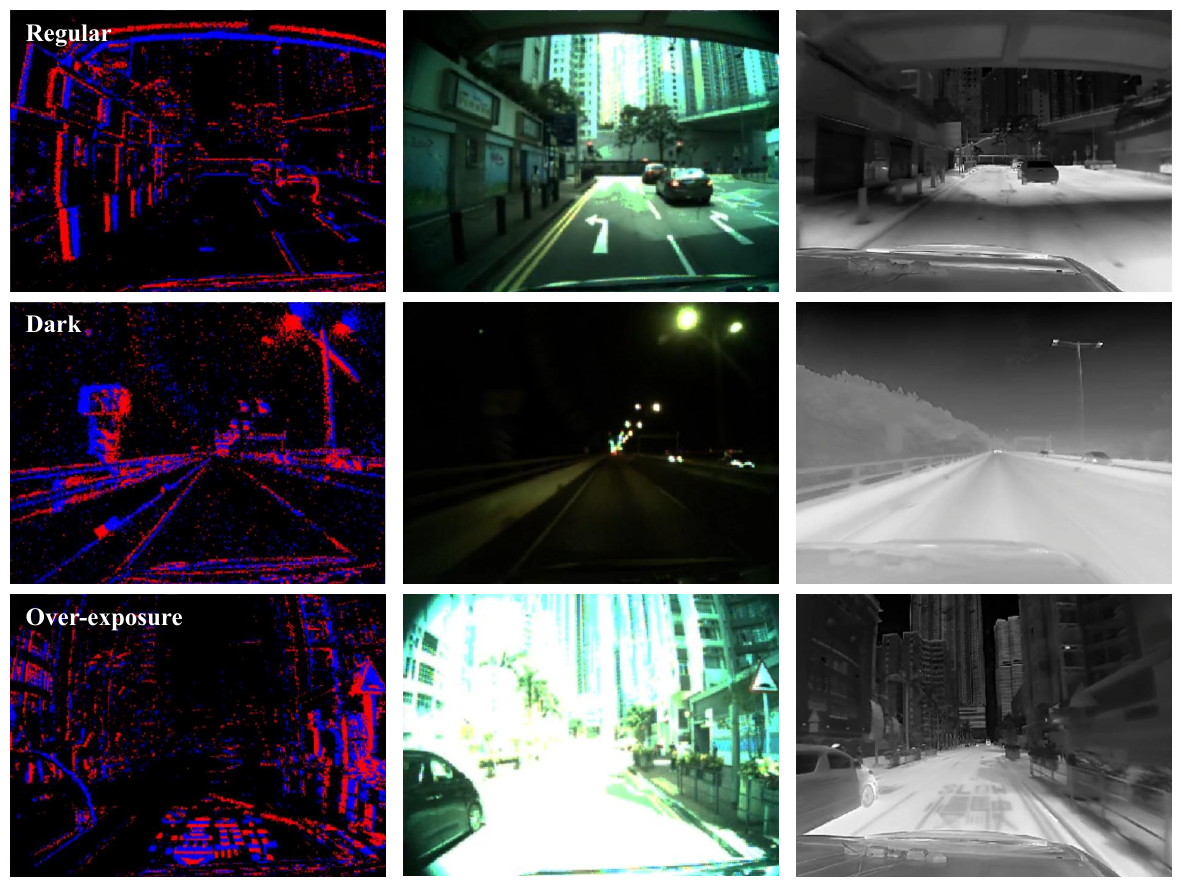}
        \caption{ Comparison of event streams, RGB images, and infrared images under different lighting conditions.}  
        \label{Comparison under different lighting conditions}
        \vspace{-1.0em}
\end{figure}

As shown in Table \ref{vision-based localization}., we evaluate the performance of VINS-MONO\cite{VINS-MONO}, ORB-SLAM3\cite{ORB-SLAM3}, and ESVIO\cite{ESVIO} across various scenes and lighting conditions on our dataset.
The accuracy is quantified using mean position error (MPE, \%), which aligns the estimated trajectory with ground-truth through 6-DOF transformation (in SE3) computed by the tool\cite{evo}.
For the VINS-Mono, we evaluate it separately using RGB images and infrared images.  
Due to the resolution provided by industrial cameras being higher in contrast to the infrared camera, we achieve superior performance when utilizing RGB images.
The ORB-SLAM3 often fails to robustly track features during high-speed vehicle movements, potentially resulting in the tracking thread restarts.
The ESVIO leverages the complementary advantages of event streams and RGB images, allowing it to handle the lack of texture in RGB images under broad illumination conditions to achieve higher accuracy.
Fig.\ref{Comparison under different lighting conditions} compares event, RGB images, and infrared images under different lighting conditions. 
The RGB images offer rich texture under regular luminance scenes in contrast to events and infrared images offer comparatively limited information, e.g., the infrared image struggles to accurately discern traffic left-turn symbol on the ground.
Conversely, the RGB image may lose numerous environmental features under the conditions of low light or over-exposure.
The infrared camera can capture infrared radiation beyond the visible spectrum and event cameras can detect pixel-level intensity changes at low latency.
Both the event camera and the infrared camera are more resilient in external varying lighting conditions, providing effective visibility compared to the industrial camera, e.g., in nighttime scenes, event cameras can capture road signs, and the infrared camera can clearly capture the surrounding bushes.


\begin{table}[htbp] 
        \renewcommand\arraystretch{1.2}
        \begin{center}
        \caption{The MPE(\%) of different visual localization algorithms on ECMD datasets}
        \label{vision-based localization}
        \resizebox{\columnwidth*1}{!}
        { 
        \begin{threeparttable}
        \begin{tabular}{c|cccc} 
        \hline  
        \multicolumn{1}{c|}{Sequence}
    & \makecell{VINS-Mono \cite{VINS-MONO} \\Mono VIO\\2.30[MP] image}
    & \makecell{VINS-Mono \cite{VINS-MONO} \\Infrared VIO\\0.33[MP] infrared}
    & \makecell{ORB-SLAM3 \cite{ORB-SLAM3} \\Stereo VO\\2.30[MP] image} 
    & \makecell{ESVIO \cite{ESVIO} \\ Stereo EVIO\\2.30[MP] image\\0.31[MP] event}\\ 
\hline
Dense street night easy a
           & \textbf{0.34}& 2.82&10.01 &0.48   \\
Urban road day easy b
           & 14.26& \textbf{7.78}& \textit{failed} & 9.63 \\
Highway day easy a
           & 1.48&2.65 & \textit{failed} &\textbf{1.04} \\
Suburban road day easy b
           & 2.36& 3.48& 9.04 & \textbf{1.35}\\
\hline        
        \end{tabular}
        \begin{tablenotes} 
        \item
        \end{tablenotes} 
        \end{threeparttable} 
        }
        \end{center}
        \vspace{-3.0em}
\end{table}

\subsection{LiDAR SLAM Evaluations}
Table \ref{LiDAR-based localization}. demonstrates the performance of LIO-SAM\cite{Lio-sam}, LVI-SAM\cite{Lvi-sam}, Fast-LIO2\cite{Fast-lio2}, Point-LIO\cite{Point-LIO} across various scenes on our dataset.
We use the same criteria introduced in Section \ref{Visual SLAM Evaluations} to evaluate the localization accuracy.
\begin{table}[htbp] 
        \vspace{-1.0em}
        \renewcommand\arraystretch{1.2}
        \begin{center}
        \caption{The MPE(\%) of different LiDAR localization algorithms on ECMD datasets}
        \label{LiDAR-based localization}
        \resizebox{\columnwidth*1}{!}
        { 
        \begin{threeparttable}
        \begin{tabular}{c|cccc} 
        \hline  
        \multicolumn{1}{c|}{Sequence}
    & \makecell{LIO-SAM \cite{Lio-sam} \\LIO}
    & \makecell{LVI-SAM \cite{Lvi-sam} \\LVIO}
    & \makecell{Fast-LIO2 \cite{Fast-lio2} \\LIO}
    & \makecell{Point-LIO \cite{Point-LIO} \\LIO}\\ 
\hline
Dense street day easy a
           & 0.25 &0.25 &\textbf{0.23} &0.25 \\
Urban road day medium
           & \textbf{0.67}& \textit{failed} &6.23 &6.34 \\
Tunnel easy a
           & \textbf{1.00}& \textit{failed}& 1.36& 20.84 \\
Highway day medium b 
           & \textbf{0.51}& \textit{failed}&0.63 &\textit{failed}  \\
Bridge day difficult a
           & \textbf{0.78}& \textit{failed}&1.33 & 15.52\\
Suburban road night medium b
           & \textbf{0.61}& \textit{failed}& 1.73& 0.72 \\
\hline        
        \end{tabular}
        \begin{tablenotes} 
        \item
        \end{tablenotes} 
        \end{threeparttable} 
        }
        \end{center}
        \vspace{-3.0em}
\end{table}

\begin{figure}[htb]  
        \centering
        \captionsetup{justification=justified}
        \includegraphics[width=1.0\columnwidth]{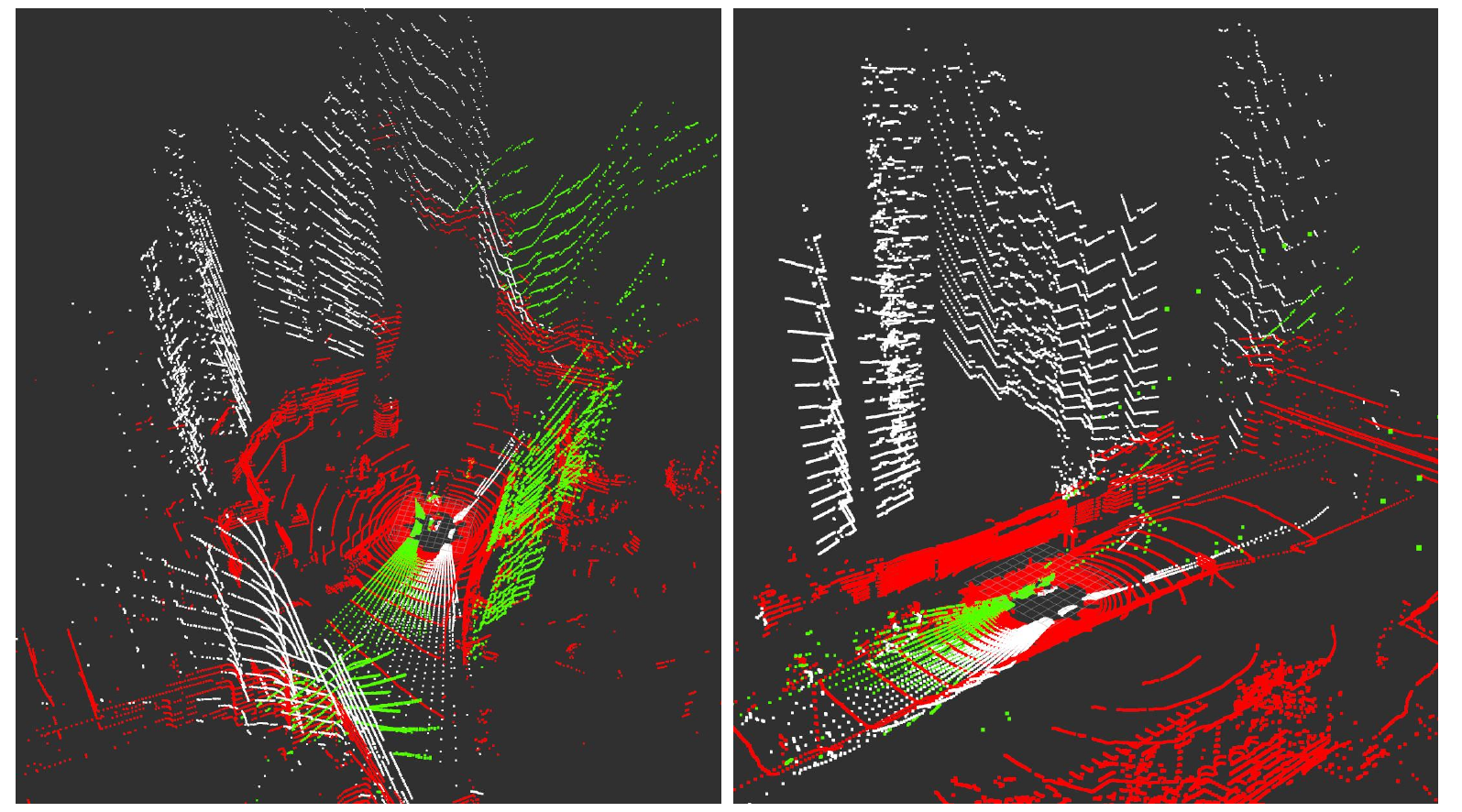}
        \caption{ Point clouds of dense street with high-rising buildings captured from three LiDARs, including two with angled installations. }  
        \label{Point clouds of dense street with high-rising buildings}
        \vspace{-1.5em}
\end{figure}

\begin{figure}[htb]  
        \centering
        \captionsetup{justification=justified}
        \includegraphics[width=1.0\columnwidth]{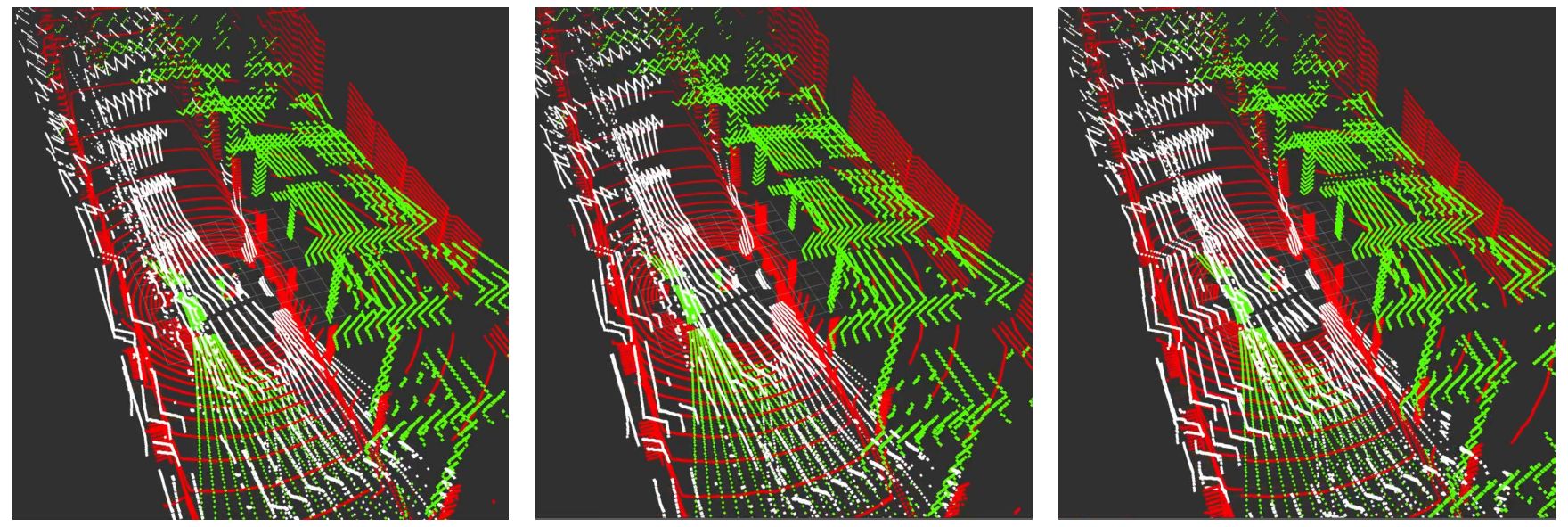}
        \caption{ LiDAR point clouds are sequentially captured in tunnel scenes. The repetitive patterns lead to the degradation challenges for localization. }  
        \label{Sequential LiDAR point cloud captures in tunnel scenes}
        \vspace{-1.5em}
\end{figure}

Due to the tilt-mounted LiDAR setups (see Fig.\ref{sensor_setup}), we are able to acquire point clouds of towering buildings situated on both sides of the street.
This installation approach compensated for the lack of vertical point clouds compared to the horizontally mounted LiDAR.
In Fig.\ref{Point clouds of dense street with high-rising buildings}, red point clouds are generated from a horizontally mounted LiDAR while white and green point clouds are generated from tilt-mounted LiDARs.
We evaluate the performance of LOAM\cite{ALOAM} using three different LiDARs (center, left, and right)in \textit{Dense\_street\_day\_medium\_circle\_a} sequences.
The MPE of LOAM using center LiDAR is 1.02\%, compared to 8.67\% using the left LiDAR and 2.00\% using the right LiDAR.
LOAM using the left LiDAR exhibits significant drift since it initially captures minimal point cloud information.
Although the LOAM merely using tilt-mounted LiDAR produces less accurate results compared to the center LiDAR, multi-LiDAR fusion can integrate complementary information, thereby improving localization accuracy and constructing more precise point cloud maps.
Meanwhile, tunnel scenes present challenges for LiDAR SLAM. 
We capture three consecutive frames of LiDAR point clouds at two-second intervals in Fig.\ref{Sequential LiDAR point cloud captures in tunnel scenes}.
It is evident that these LiDAR point clouds exhibit high similarity in the tunnel environment, potentially resulting in degradation phenomena and inaccurate state estimation.

\section{ISSUES} 
\label{ISSUES}



\subsection{LiDAR-Camera Interference}
Due to space limitations, we positioned LiDAR closer to the event cameras. 
As a consequence, the infrared wavelengths emitted by LiDAR directly impinge on the photoreceptor of event cameras, resulting in continuous disturbances and flickering in the captured images and event streams. 
To address this issue, we implement infrared filters on event cameras to counteract the effect. 
However, this intervention led to a compromise, resulting in a degradation of the quality of the recorded event data.

\subsection{Event Artifacts}
During the night or low illumination scenarios, we observed that when event cameras were directly toward a glowing light source, such as street lights or store lighting, event streams would exhibit persistent flickering and produce artifacts around the light source.
This could potentially lead to a distorted view of the observed object.
We postulate this phenomenon is related to the inherent principle of event cameras, and presently, there is no known solution to address this issue.


\section{CONCLUSIONS} 
\label{CONCLUSIONS}

In this paper, we propose an event-centric autonomous driving dataset generated with multiple sensors across various scenarios for developing SLAM algorithms.
All sensors undergo meticulous calibration and are temporally synchronized at the hardware level.
We employ the GNSS-RTK/INS navigation system, which provides centimeter-level accuracy, to acquire precise ground-truth of the vehicle.
Furthermore, we conduct the evaluation of various state-of-the-art visual and LiDAR SLAM algorithms while identifying their constraints.
We hope this dataset could contribute to the development of visual and LiDAR SLAM.
In future work, we intend to expand the dataset to encompass additional tasks, including semantics, optical flow, depth estimation, etc.





\bibliographystyle{IEEEtran} 
\bibliography{references.bib} 



\vspace{-3.0em}
\begin{IEEEbiography}[{\includegraphics[width=1in,height=1.25in,clip,keepaspectratio]{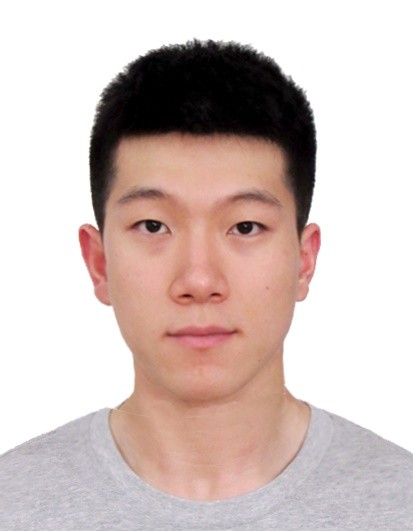}}]{Chen Peiyu}
received the BSc degrees in automation from the Nanjing University of Science and Technology, China, and MSc degrees in computer control \& automation from the Nanyang Technological University, Singapore, in 2020 and 2022, respectively. 
He is currently working forward the Ph.D. degree at the University of Hong Kong.  
His research interests include robotics, visual-inertial simultaneous localization and mapping, nonlinear control, and so on.
\end{IEEEbiography}

\begin{IEEEbiography}[{\includegraphics[width=1in,height=1.25in,clip,keepaspectratio]{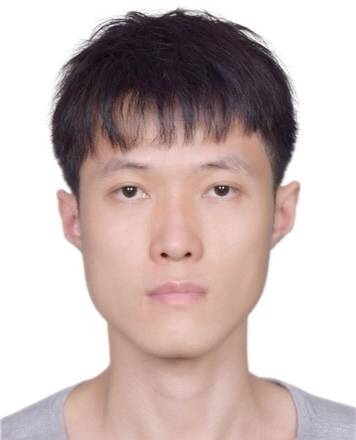}}]{Guan Weipeng}
obtained his Bachelor degree in Electronic Science \& Technology, as well as Master degree in Control Theory and Control Engineering from the South China University of Technology. 
He is currently pursuing his PhD degree in Robotics at the University of Hong Kong. 
He has worked with several reputable organizations, including: Samsung Electronics, Huawei Technologies, The Chinese Academy of Sciences, The Chinese University of Hong Kong, The Hong Kong University of Science and Technology, etc.  
He has also served as a technical consultant for multiple companies, such as TCL.
Moreover, he has published over 60 research articles in prestigious international journals and conferences, as well as holds more than 40 authorized patents. 
His research interests primarily focus on robotics, event-based VO/VIO/SLAM, visible light positioning, etc.
\end{IEEEbiography}

\begin{IEEEbiography}[{\includegraphics[width=1in,height=1.25in,clip,keepaspectratio]{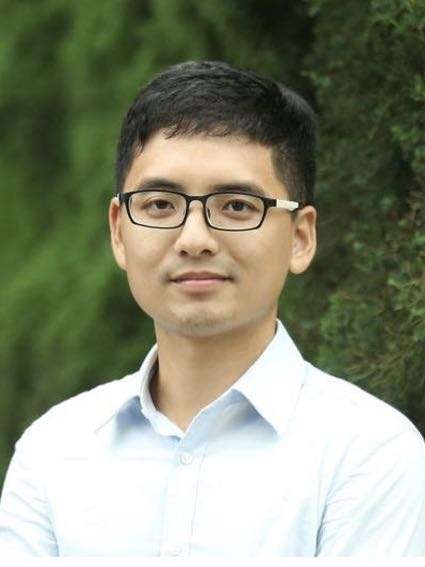}}]{Huang Feng}
received his bachelor’s degree from Shenzhen University in Automation in 2014 and MSc in Electronic Engineering at Hong Kong University of Science and Technology in 2016.
He is a Ph.D. student in the Department of Aeronautical and Aviation Engineering, Hong Kong Polytechnic University. 
His research interests including localization and sensor fusion for autonomous driving.
\end{IEEEbiography}

\begin{IEEEbiography}[{\includegraphics[width=1in,height=1.25in,clip,keepaspectratio]{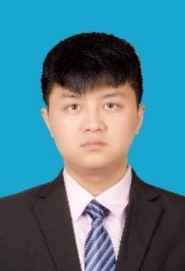}}]{Zhong Yihan}
obtained his bachelor's degree in process equipment and control engineering from Guangxi University in 2020 and a Master's degree from The Hong Kong Polytechnic University (PolyU). 
He is currently a Ph.D. student at the Department of Aeronautical and Aviation Engineering (AAE) of PolyU. 
His research interests include factor graph optimization-based collaborative positioning and low-cost localization.
\end{IEEEbiography}

\vspace{6.0em}
\begin{IEEEbiography}[{\includegraphics[width=1in,height=1.25in,clip,keepaspectratio]{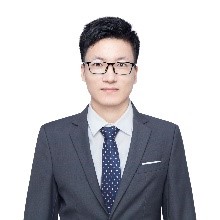}}]{Wen Weisong}
(Member, IEEE) received a BEng degree in Mechanical Engineering from Beijing Information Science and Technology University (BISTU), Beijing, China, in 2015, and an MEng degree in Mechanical Engineering from the China Agricultural University, in 2017. After that, he received a PhD degree in Mechanical Engineering from The Hong Kong Polytechnic University (PolyU), in 2020. He was also a visiting PhD student with the Faculty of Engineering, University of California, Berkeley (UC Berkeley) in 2018. Before joining PolyU as an Assistant Professor in 2023, he was a Research Assistant Professor at AAE of PolyU since 2021. He has published 30 SCI papers and 40 conference papers in the field of GNSS (ION GNSS+) and navigation for Robotic systems (IEEE ICRA, IEEE ITSC), such as autonomous driving vehicles. He won the innovation award from TechConnect 2021, the Best Presentation Award from the Institute of Navigation (ION) in 2020, and the First Prize in Hong Kong Section in Qianhai-Guangdong-Macao Youth Innovation and Entrepreneurship Competition in 2019 based on his research achievements in 3D LiDAR aided GNSS positioning for robotics navigation in urban canyons. The developed 3D LiDAR-aided GNSS positioning method has been reported by top magazines such as Inside GNSS and has attracted industry recognition with remarkable knowledge transfer. 
\end{IEEEbiography}

\begin{IEEEbiography}[{\includegraphics[width=1in,height=1.25in,clip,keepaspectratio]{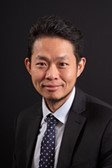}}]{Hsu Li-Ta}
 received B.S. and Ph.D. degrees in aeronautics and astronautics from National Cheng Kung University, Taiwan, in 2007 and 2013, respectively. He is currently an associate professor with the Department of Aeronautical and Aviation Engineering. The Hong Kong Polytechnic University, before he served as a post-doctoral researcher at the Institute of Industrial Science at the University of Tokyo, Japan. In 2012, he was a visiting scholar at University College London, the U.K. His research interests include GNSS positioning in challenging environments and localization for pedestrians, autonomous driving vehicle, and unmanned aerial vehicle.
\end{IEEEbiography}

\begin{IEEEbiography}[{\includegraphics[width=1in,height=1.25in,clip,keepaspectratio]{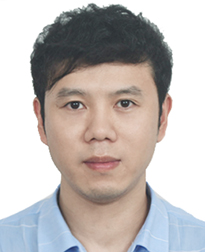}}]{Lu Peng}
obtained his BSc degree in automatic control and MSc degree in nonlinear flight control both from Northwestern Polytechnical University (NPU). He continued his journey on flight control at Delft University of Technology (TU Delft) where he received his PhD degree in 2016. After that, he shifted a bit from flight control and started to explore control for ground/construction robotics at ETH Zurich (ADRL lab) as a Postdoc researcher in 2016. He also had a short but nice journey at University of Zurich \& ETH Zurich (RPG group) where he was working on vision-based control for UAVs as a Postdoc researcher. He was an assistant professor in autonomous UAVs and robotics at Hong Kong Polytechnic University prior to joining the University of Hong Kong in 2020.

Prof. Lu has received several awards such as 3rd place in 2019 IROS autonomous drone racing competition and best graduate student paper finalist in AIAA GNC (top conference in aerospace). He serves as an associate editor for 2020 IROS (top conference in robotics) and session chair/co-chair for conferences like IROS and AIAA GNC for several times. He also gave a number of invited/keynote speeches at multiple conferences, universities and research institutes.
\end{IEEEbiography}


\end{document}